\documentclass[letterpaper, 10 pt, conference]{ieeeconf}  

\IEEEoverridecommandlockouts                              

\usepackage{hyperref}
\usepackage{cite}
\usepackage[cmex10]{amsmath}
\usepackage{algorithmic}
\usepackage{array}
\usepackage{float}
\usepackage{graphicx} 
\usepackage{epstopdf}
\usepackage{mathptmx} 
\usepackage{times} 
\usepackage{amsmath} 
\usepackage{amssymb}  
\usepackage{dblfloatfix}
\usepackage{url}
\usepackage{subcaption}
\usepackage{lipsum}
\usepackage{adjustbox,lipsum}
\usepackage{booktabs}
\usepackage{caption}
\usepackage{multirow}
\usepackage{tabularx,ragged2e,booktabs,caption}
\usepackage{textcomp}
\usepackage{hyphenat}
\usepackage{diagbox}
\usepackage{mathtools}
\usepackage{booktabs}
\usepackage{threeparttable}

\title{\LARGE \bf LSTM-based Network for Human Gait Stability Prediction \\ in an Intelligent Robotic Rollator}

\author{Georgia Chalvatzaki, Petros Koutras, Jack Hadfield,\\ Xanthi S. Papageorgiou, Costas S. Tzafestas and Petros Maragos\\
School of Electrical and Computer Engineering, National Technical University of Athens, Greece\\
{\tt\small \{gchal, xpapag\}@mail.ntua.gr},
{\tt\small \{pkoutras, ktzaf, maragos\}@cs.ntua.gr}
\thanks{This research has been co‐financed by the European Union and Greek national funds through the Operational Program Competitiveness, Entrepreneurship and Innovation, under the call RESEARCH – CREATE – INNOVATE  (project code:T1EDK- 01248 / MIS: 5030856).}
}

\begin{document}

\maketitle
\thispagestyle{empty}
\pagestyle{empty}

\begin{abstract}
In this work, we present a novel framework for on-line human gait stability prediction of the elderly users of an intelligent robotic rollator using Long Short Term Memory (LSTM) networks, fusing multimodal RGB-D and Laser Range Finder (LRF) data from non-wearable sensors. A Deep Learning (DL) based approach is used for the upper body pose estimation. The detected pose is used for estimating the body Center of Mass (CoM) using Unscented Kalman Filter (UKF). An Augmented Gait State Estimation framework exploits the LRF data to estimate the legs' positions and the respective gait phase. These estimates are the inputs of an encoder-decoder sequence to sequence model which predicts the gait stability state as Safe or Fall Risk walking. It is validated with data from real patients, by exploring different network architectures, hyperparameter settings and by comparing the proposed method with other baselines. The presented LSTM-based human gait stability predictor is shown to provide robust predictions of the human stability state, and thus has the potential to be integrated into a general user-adaptive control architecture as a fall-risk alarm. 
\end{abstract}

\vspace{-.2cm}

\section{Introduction}

\subsection{Motivation}
The worldwide population aged over 65 rises exponentially according to recent reports of the United Nations \cite{UNpop}. 
Amongst others the mobility problems prevail in the elder society. Ageing and many pathologies invoke changes in walking speed and stability \cite{Hausdorff2007555}, while 30\% of the aged population is reported to have fallen every year. Especially, changes in gait speed are related to the functional independence and mobility impairment of the elderly \cite{Pinillos16}, and are closely connected to fall incidents. 

In the last 15 years robotics research has focused on robotic mobility assistive devices, aiming to provide postural support and walking assistance \cite{pamm,graf04,guido05,iwalker08,JaistROMAN14}, as well as sensorial and cognitive assistance to the elderly \cite{Jenkins2015}.  
Their goal is to increase the user mobility, while avoiding the anxiety and frustration imposed by the common walking aids. In our previous work \cite{gchal_iros17, gchal_iros18}, we have shown that for a robotic rollator, that aims to support patients of different mobility status, user-adaptation is important.
\begin{figure}[htpb]
      \centering
      \includegraphics[width=0.45\textwidth]{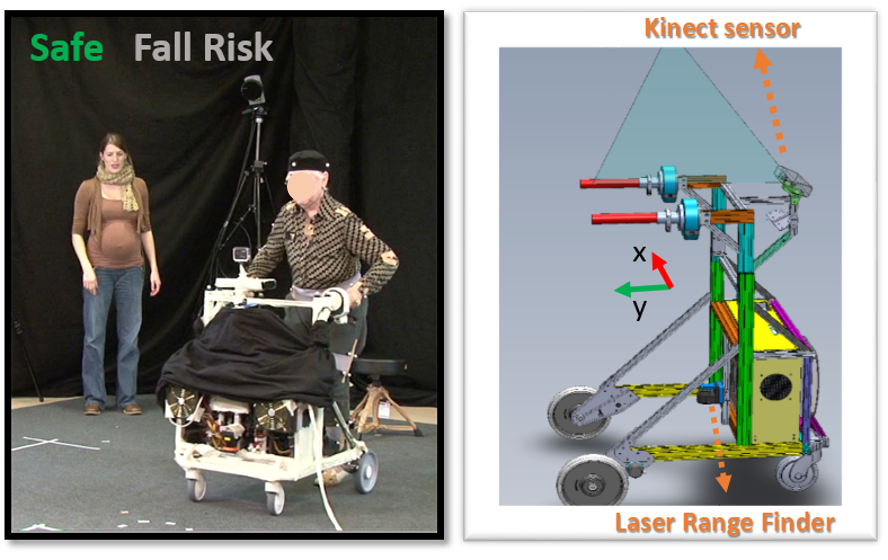}
      \caption{\small \textbf{Left}: Elderly patient walking supported by a robotic rollator. The predicted states for Gait Stability may be Safe walking or Fall Risk. \textbf{Right}: a CAD drawing of the rollator with RGB-D and LRF sensors.}       
      \label{fig:motivation}
      \vspace{-0.78cm}
\end{figure}
Specifically, a handy system should be able to assess the mobility state of the user and adapt the control strategies accordingly, and also monitor the rehabilitation progress and provide fall prevention.
 Although computing various gait parameters and the on-line classification of the pathological gait status of the user plays a significant role in the adaptation of a context-aware controller, there are also issues of gait stability that should be addressed.

The purpose of this paper is to present, analyse and evaluate a novel and robust method for on-line gait stability analysis of elderly subjects walking with a mobility assistant platform (Fig. \ref{fig:motivation}), fusing the information from an RGB-D camera that captures the upper body and a LRF monitoring the legs motion in the sagittal plane. Specifically, we use a DL approach for detecting the upper body pose and track the respective CoM through time along with an augmented gait state estimation from the LRF \cite{gchal_ral18}. We propose a LSTM-based network for predicting the stability state of the patient, by classifying his gait as safe or fall-risk at each instance. We present a new method for predicting the gait stability state of robotic rollator users with variable pathological gait conditions, using only non-wearable sensors, in order to supply the robotic system with an alarm regarding fall risk episodes. The main goal is to integrate this information into a user-adaptive context-aware robot control architecture for the robotic assistant platform that would prevent possible falls.

\subsection{Related Work}
Fall detection and prevention is a hot topic in the field of assistive robotics \cite{s141019806}.
Most of the proposed control strategies for robotic assistive platforms in literature do not deal with the problem of fall prevention and research works focus on navigation and obstacle avoidance \cite{Chuy08,Cifuentes2016_1,Geravand2016}. However, there exist some targeted research focusing on incorporating strategies for preventing or detecting fall incidents and facilitating user's mobility. In \cite{Hirata06, Hirata08} the authors developed an admittance controller for a passive walker with a fall-prevention function considering the position and velocity of the user, utilizing data from two LRFs. They model the user as a solid body link, in order to compute the position of the center of gravity \cite{Takeda17}, based on which they applied a braking force on the rollator to prevent falls. A fall detection for a cane robot was presented in \cite{Di2013,Di2016}, that computes the zero moment point stability of the elderly, using on-shoe sensors that provide ground force reactions. 

Regarding the extraction of gait motions, different types of sensors have been used \cite{Bae2011961,accelom}. 
Gait analysis can be achieved by using Hidden Markov Models for
modelling normal \cite{xpapag14} and pathological human gait \cite{xpapag15}, and extracting gait parameters \cite{xpapag16Biorob}. Recently, we have developed a new method for online augmented human state estimation, that uses Interacting Multiple Model Particle Filters with Probabilistic Data Association \cite{gchal_ral18}, which tracks the users' legs using data from a LRF, while it provides real-time gait phase estimation. We have also presented a new human-robot formation controller that utilizes the gait status characterization for user adaptation towards a fall preventing system \cite{gchal_iros18}.

Gait stability is mostly analysed by using wearable sensors \cite{7523760}, like motion markers placed on the human body to calculate the body's CoM and the foot placements \cite{LUGADE2011406}, and force sensors to estimate the center of pressure of the feet \cite{s130810151}. Gait stability analysis for walking aid users can be found in \cite{COSTAMAGNA2017167}. Regarding stability classification, an early approach can be found in \cite{Taghvaei17}, where the authors use the body skeleton provided by the RGB-D Kinect sensor as input and perform action classification to detect four classes of falling scenarios. However, the system was tested only with a  physical  therapist performing different walking problems.

Human pose estimation is a challenging topic due to the variable formations of the human body, the parts occlusions, etc. The rise of powerful DL frameworks along with the use of large annotated datasets opened a new era of research for optimal human pose estimation \cite{Andriluka14}. Most approaches provide solutions regarding the detection of the 2D pose from color images by detecting keypoints or parts on the human body \cite{Wei2016,Cao2017} achieving high accuracy. The problem of 3D pose estimation is more challenging \cite{martinez_2017_3dbaseline}, as the detected poses are scaled and normalized. Recent approaches aim to solve the ambiguity of 2D-to-3D correspondences by learning 3D poses from single color images \cite{pavlakos2018humanshape, pavlakos2018ordinal}. Another relevant research topic concerns the tracking of human poses  \cite{girdhar2018detecttrack}, but while they achieve improved levels of accuracy compared to previous methods, due to the contribution of DL, the high estimation error makes it prohibitive to integrate it into a robotic application that requires high accuracy and robustness. A recent application of pose estimation for a robotic application can be found in \cite{zw2018}.

Our contributions presented in this work is the design of a novel deep-based framework for on-line human gait stability state prediction using the detection of the upper body 3D pose and the respective CoM estimation from an RGB-D camera and the human gait states  estimated from LRF data. Differing from the common gait stability analysis methods in literature, we propose an LSTM-based network for fusing the multi-modal information, in order to decode the hidden interaction of the body's CoM with the legs motion and the gait phases, in order to predict the gait stability of the elderly considering two possible classes: safe and fall-risk walking. The proposed on-line LSTM-based human gait stability predictor is evaluated using multi-modal data from real patients. To justify the model selection, we present an exploratory study regarding the network architecture, the selected hyperparameters and compare the performance of our framework with baseline methods. The results demonstrate the great efficiency of the LSTM-based approach to provide robust predictions of the human stability state, and show its potential to be integrated into a general user-adaptive control architecture as a fall-risk alarm. 
\begin{figure*}[ht!]
      \centering
      \includegraphics[width=0.85\textwidth]{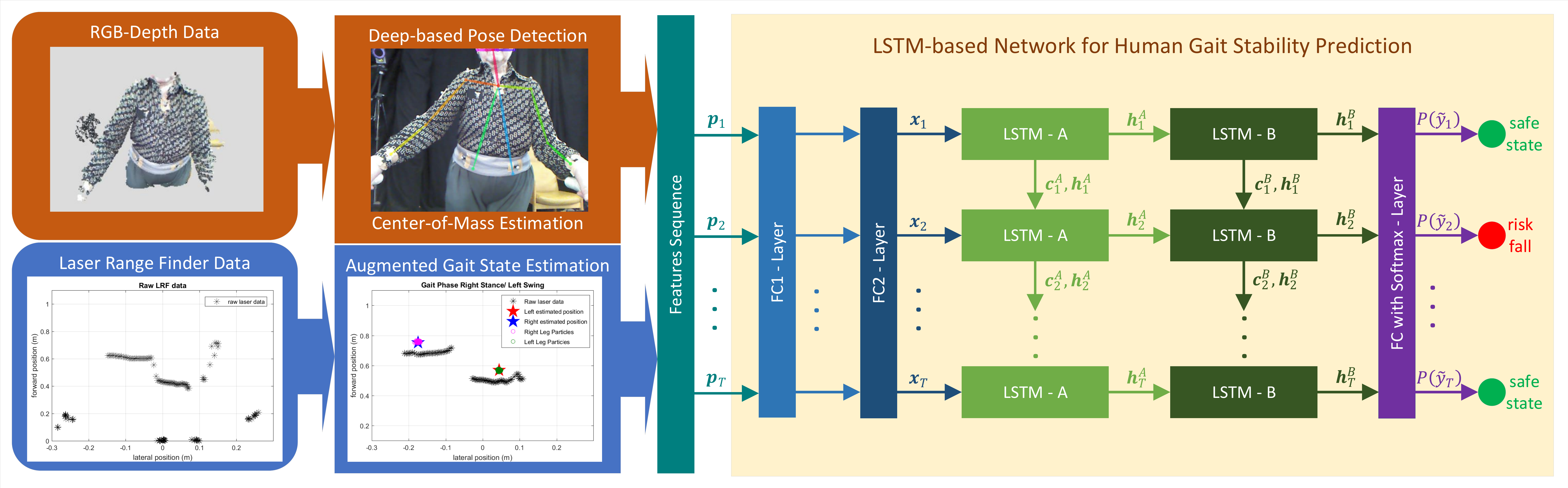}
      \caption{\small Overview of the proposed LSTM-based On-line Human Gait Stability Prediction framework.}       
      \label{fig:icra_lstm}
      \vspace{-0.65cm}
\end{figure*}
      \vspace{-0.2cm}

\section{Human Gait Stability}\label{gaitStabdescr}
During walking the body is in a continuous state of instability \cite{Bruijn20170816}. Gait stability is described by the interaction of the position and velocity of the CoM w.r.t. the Base of Support (BoS) in the horizontal plane. The BoS is the imaginary parallelogram formed by the contact of at least one foot with the ground. During double support phases the BoS covers its largest area, while in single leg support phases (one foot in stance phase while the other swings through) the BoS covers smaller areas. Biomechanics considers an inverted pendulum model to describe the CoM-BoS interaction \cite{Pai1997CenterOM}.

When the projection of the CoM lies inside the BoS the body is stable. However, during gait the CoM is outside the BoS for the single leg support phases, for the most time in a gait cycle. Each foot contact, initiating a new gait cycle, prevents a potential fall \cite{LUGADE2011406}. One indicator of human stability is the distance of the CoM to the boundaries of the BoS, which is also the stability measure used in this study.

When the CoM is inside the BoS their respective distance is called stability margin, and when the CoM lies outside the BoS the distance is called CoM separation. The measures of these distances are indicative of the stability of a person while walking. Although a human supported by a walking aid has an enlarged BoS, the reported high fall incidents of walking aid users \cite{Mettelinge2015UnderstandingTR}, along with the fact that users often disengage their hands from the aid to perform several actions, led us to consider the general notion of the human-centric BoS in this particular paper.

\section{Method}
In Fig. \ref{fig:icra_lstm} an overview of the proposed LSTM-based human gait stability prediction framework is presented. The proposed method uses multimodal RGB-D and LRF data. The RGB-D data are employed for a deep-based pose detection and the estimation of the CoM position. The LRF data are used in an augmented gait state estimation framework for estimating the legs' position and the gait phase. These estimated human-motion related parameters constitute the features of the LSTM-based Network for predicting the gait stability state at each time instance, using a binary description of the user's state as ``Safe" or ``Fall Risk" state. The components of the whole framework are described below. 
\vspace{-0.35cm}

\subsection{Augmented Gait State Estimation}
The augmented human gait state estimation was proposed in \cite{gchal_ral18}. It is a novel framework for efficient and robust leg tracking from LRF data along with human gait phases predictions (Fig. \ref{fig:icra_lstm}). Our approach uses two Particle Filters (PFs) and Probabilistic Data Association (PDA) with an Interacting Multiple Model (IMM) scheme for a real-time
selection of the appropriate motion model of the PFs. The IMM is a first-order Markov model with the following gait states: a) Left Double Support (DS), b) Left Stance/ Right Swing, c) Right DS, d) Right Stance/ Left Swing. Each state refers to both legs and imposes a different motion model for the PFs. The legs are tracked, and a maximum likelihood estimation predicts the current gait phase.
A thorough analysis of the methodology is provided in \cite{gchal_ral18}. From this method the augmented gait state $ \mathbf{s}^{gait}_t= \left[ {\begin{array}{*{20}{c}}
x^l & y^l & x^r & y^r & phase
\end{array}} \right]^T$, consisting of the estimated legs' positions along the axes (\textit{l,r} refer to left and right leg respectively) and the respective gait phase, is used in the LSTM-based network (Fig. \ref{fig:icra_lstm}).
\vspace{-0.15cm}

\subsection{Deep-based Pose Detection and CoM Estimation}
For the upper body pose estimation we employ the RGB images and the respective depth maps provided by the Kinect sensor, which is mounted on the rollator (Fig. \ref{fig:motivation}). The 2D positions of the keypoints are detected using the Open Pose (OP) Library \cite{Cao2017} with fixed weights. OP uses a bottom-up representation of associations of the locations and orientations of the limbs through the images and a two branch  multi-stage convolutional neural network to predict the keypoints' 2D positions. The third dimension of the keypoints is obtained by depth maps. The depth maps need to be transformed to the image plane using the calibration matrix of the camera. For this purpose we have applied the method of \cite{zw2018}. 

Despite the high performance of the OP framework, the close proximity of the human body to the Kinect sensor, the occlusions of body parts, such as the head or large part of the arms, from the camera's field of view (Fig. \ref{fig:icra_lstm}), or even the high reflectivity due to ambient light, lead to many detection losses and misdetections. Thus, tracking is required. When the pose is detected, we use the torso keypoints to compute the 3D position of the torso center, as the median of the keypoints. From the experimental analysis of the motion markers, which are used as Ground Truth (GT), we have modelled the upper body as ellipsoid with its center being the CoM. We use statistics about the ellipsoid fitting to translate the torso center in a position representing the CoM. Finally, the detected CoM is transformed from the camera coordinate frame to the robot frame.

The detected CoM positions are the observations of an UKF that tracks and estimates the CoM state through time. The UKF servers many purposes; it is used to model and predict the nonlinear CoM motion \cite{GaitAnalPerry92, 14-georgia}, it filters the noisy observations, it compensates the different frame rates between the Kinect and the LRF sensor by giving predictions of the CoM states during the periods when the sensor does not transmit measurements, and also provides estimates when we do not get/accept the corrupted keypoints detections. 

The tracking framework employs the well-known prediction and update equations of the UKF, described in \cite{UKF2000}. The CoM motion model includes the following kinematic equations: 
\vspace{-0.15cm}
{\small
\begin{equation}
\begin{array}{l}
q^x_t = q^x_{t - 1} + \upsilon_{t - 1} \cdot \cos (\omega^z_{t-1}\cdot \Delta t) \cdot \Delta t\\
q^y_t = q^y_{t - 1} + \upsilon_{t - 1} \cdot \sin (\omega^z_{t - 1}\cdot \Delta t) \cdot \Delta t\\
\upsilon_t = \upsilon_{t - 1} + \eta_{\upsilon}\\
\omega ^z_t = \omega^z_{t - 1} + \eta_{{\omega^z}}
\end{array}
\end{equation}}
\normalsize
where $t$ is the discrete time, $q^x_t,q^y_t$ the position along the axes, $\upsilon_t,\omega ^z_t$ the linear and angular velocities, $\Delta t$ is the time interval in which we make predictions and $\eta_{\upsilon},\eta_{\omega_z}$ are the linear and angular velocity noises modelled as zero-mean white Gaussian with standard deviations $\sigma_{\upsilon}=0.98 \mbox{ m/sec} \mbox{ and } \sigma_{\omega_z}=1.88 \mbox{ rad/sec}$ respectively, computed from the motion markers data.

The linear observation model in UKF considers only the detected positions $q^x_t,q^y_t$ of the CoM. In the observation model  the noise vectors are modelled as white Gaussians with standard deviations $0.15$ m for the $q^x$ variable and $0.2$ m for the $q^y$ variable, learned from experiments (higher variability in measuring the $q^y$ from the camera's depth map). Only the estimated CoM position from the UKF
 is fed to the LSTM-based network (Fig. \ref{fig:icra_lstm}); let us denote it as $\mathbf{s}^{CoM}_t={\left[ {\begin{array}{*{20}{c}}
{{q^x}}&{{q^y}}
\end{array}} \right]^T}$.
\vspace{-0.2cm}
\begin{figure}[t!]
      \centering
      \includegraphics[width=0.35\textwidth,trim={1.2cm 0cm 1.2cm 0cm },clip]{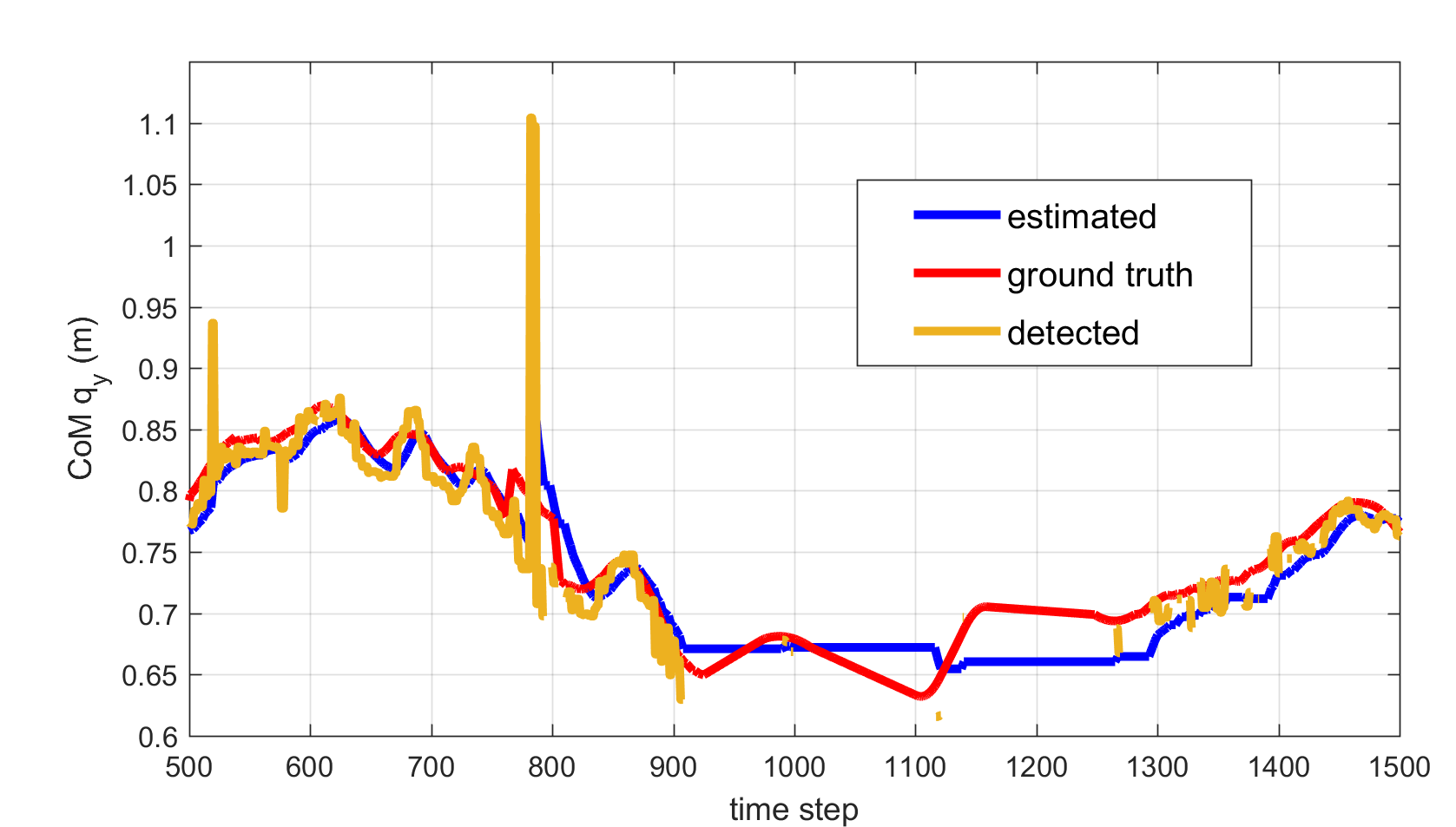}
      \caption{\small CoM forward displacement.}\label{fig:CoM_estim1}
      
\vspace{-.55cm}
\end{figure}

\subsection{LSTM-based Network for Gait Stability Prediction}

In our learning based method for gait stability prediction we employ a Neural Network (NN) architecture based on LSTM units \cite{hochreiter1997long}. LSTM constitutes a recurrent NN that can effectively learn long-term dependencies by incorporating memory cells $\mathbf{c}_t$ that allow the network to learn when to forget previous hidden states and when to update hidden states given new information. The overall architecture is an encoder-decoder sequence-to-sequence model, considering only past input vectors to make predictions. It consists of two Fully Connected (FC) layers, two LSTM layers and a last FC layer followed by Softmax (Fig.~\ref{fig:icra_lstm}).  

\noindent \textbf{Input representation:} Let $\mathbf{p}_t=\left[ \begin{array}{*{20}{c}}
\mathbf{ s}^{CoM}_t & \mathbf{\ s}^{gait}_t \end{array} \right]$, be the standardised observations (with zero mean and unit variance) at each time instant \textit{t} and $\{\mathbf{p}_t\}_{1}^T$ the sequence of our observations in a temporal window of length $T$. These observations are transformed to the LSTM inputs $\{\mathbf{x}_t\}_{1}^T$ by feeding them to a network with two FC layers:
\begin{equation}
\small
\mathbf{x}_t = \rho(\mathbf{W}_{FC2}\cdot \rho( \mathbf{W}_{FC1} \cdot\mathbf{p}_t + \mathbf{b}_{FC1}) + \mathbf{b}_{FC2}),
\end{equation}  
where $\mathbf{W}_{FC1}, \mathbf{W}_{FC2}$ and $\mathbf{b}_{FC1}, \mathbf{b}_{FC2}$ are the weight and the biases of the two linear layers and $\rho( \cdot )$ is a Rectified Linear Unit (ReLU) nonlinearity, defined as $\rho(\mu) = max(0,\mu)$.  These two FC layers have the role of an encoder of the features, helping to encode the nonlinear function of the CoM with the legs for detecting instability. In this way, we learn a static transformation and find a better representation for the observations before feeding them to the LSTM unit that models time dependencies.

\noindent \textbf{LSTM Unit:} LSTM is composed of an input gate $\mathbf{i}$, an input modulation gate $\mathbf{g}$, a  memory cell $\mathbf{c}$,  a forget gate $\mathbf{f}$ and an output gate $\mathbf{o}$. LSTM takes the computed inputs $\mathbf{x}_t$ at each time step $t$, the previous estimations for the hidden state $\mathbf{h}_{t-1}$ and the memory cell state $\mathbf{c}_{t-1}$, in order to update their states using the equations:

\vspace{-0.3cm}
{\small
\begin{eqnarray}
\mathbf{i}_t &=& \sigma(\mathbf{W}_{xi}\mathbf{x}_t + \mathbf{W}_{hi}\mathbf{h}_{t-1} + \mathbf{b}_i) \nonumber \\
\mathbf{f}_t &=& \sigma(\mathbf{W}_{xf}\mathbf{x}_t + \mathbf{W}_{hf}\mathbf{h}_{t-1} + \mathbf{b}_f) \nonumber \\
\mathbf{o}_t &=& \sigma(\mathbf{W}_{xo}\mathbf{x}_t + \mathbf{W}_{ho}\mathbf{h}_{t-1} + \mathbf{b}_o)  \\
\mathbf{g}_t &=& \phi(\mathbf{W}_{xg}\mathbf{x}_t + \mathbf{W}_{hg}\mathbf{h}_{t-1} + \mathbf{b}_g)  \nonumber \\
\mathbf{c}_t &=& \mathbf{f}_t \odot   \mathbf{c}_{t-1} +  \mathbf{i}_t \odot  \mathbf{g}_t  \nonumber \\
\mathbf{h}_t &=& \mathbf{o}_t \cdot \phi(\mathbf{c}_t)  \nonumber
\end{eqnarray}}
\normalsize
 \begin{table}[t!]
 \caption{\small{Demographics}}\label{tab:demograph}
\centering
    \resizebox{0.35\textwidth}{!}{%
\begin{tabular}{c|c|c|c|c}\toprule
\hline
subject & Sex &Age& Height (cm)& Weight (kg)\\ \hline
\midrule
   1 & F & 80& 153,5 &64,1\\ \hline
   2 & F & 77& 164 & 89.5\\ \hline
   3 & F & 80 & 140,5 & 73,1 \\ \hline
   4 & M & 85 & 170 & 75 \\ \hline
   5 & M & 81 & 178 & 61,4 \\ \hline
     \bottomrule
\end{tabular}}
\vspace{+0.15cm}
\caption{\small{Average RMSE for the CoM estimation}}\label{tab:com_rmse}
\centering
    \resizebox{0.40\textwidth}{!}{%
\begin{tabular}{cccccc|c}\toprule
\hline
subject & 1 &2& 3& 4 & 5 & mean\\ \hline
\midrule
   average RMSE (cm) & 2,65 & 6,13& 1,2 &6,09& 2,5 & 3,71\\ 
     \bottomrule
\end{tabular}}
\vspace{-.6cm}
\end{table}
where we have the equations of the four gates: input gate $\mathbf{i}_t \in \mathbb{R}^N$, forget gate $\mathbf{f}_t \in  \mathbb{R}^N$, output gate $\mathbf{o}_t  \in \mathbb{R}^N$, input modulation gate $\mathbf{g}_t  \in \mathbb{R}^N$) that modulates the memory cell $\mathbf{c}_t  \in  \mathbb{R}^N$ and the hidden state $\mathbf{h}_t  \in  \mathbb{R}^N$ with $N$ hidden units. 
Symbol $\odot$ represents element-wise multiplication, the function $\sigma: \mathbb{R}\rightarrow[0,1], \mbox{ } \sigma(\mu)=\frac{1}{1+e^{-\mu}}$ is the sigmoid non-linearity and
$\phi: \mathbb{R}\rightarrow[-1,1], \phi(\mu)=\frac{e^\mu-e^{-\mu}}{e^\mu+e^{-\mu}}$ is the hyperbolic tangent non-linearity. $\mathbf{W}_{xn},\mathbf{W}_{hn}$, with $n=\{i,f,o,g\}$ are the weight matrices of the input and recurrent connection of each gate and $\mathbf{b}_n$ denotes the bias vectors for each gate. The parameters of the $\mathbf{W}_{xn},\mathbf{W}_{hn}$ and $\mathbf{b}_n$ are learned during the training of the NN.

In the last layer (Fig. \ref{fig:icra_lstm}),  we estimate the classes of gait stability $y_t$ at each time step $t$, $y_t \in [ 0,1] $ with ``0" being the Safe class and ``1" the Fall Risk class, by learning a linear transformation from the hidden states $\mathbf{h}_t$ to the output state $\mathbf{\tilde{y}}_t$, described by: $\mathbf{\tilde{y}}_t=\mathbf{W}_{hy}\mathbf{h}_t + \mathbf{b}_y$, where again $\mathbf{W}_{hy}$ is the weight matrix and $\mathbf{b}_y$ the bias vector of the output layer. Then, the 
probability of having a Fall Risk gait is given by taking the softmax: 

\vspace{-0.3cm}
{\small
\begin{equation}
P(y_t=1| \{\mathbf{p}_\tau\}_{1}^t; \mathbf{W}, \mathbf{b})  = \frac{e^{\mathbf{\tilde{y}}_{t,1}}}{e^{\mathbf{\tilde{y}}_{t,0}} + e^{\mathbf{\tilde{y}}_{t,1}}},  \label{eq:softmax}
\end{equation}}
where $\mathbf{W}, \mathbf{b}$ denote the trainable parameters of the whole network and $\{\mathbf{p}_\tau\}_{1}^t$ are the observations until time $t$.
\begin{table*}[htpb]
\caption{Performance results of different architectures of LSTM-based Prediction Models} \label{tab:lstms_exam}
\centering
 \begin{threeparttable}[b]
    \resizebox{\textwidth}{!}{%
    \large
\begin{tabular}{@{}lcccccccccccccccccccc@{}}\toprule
\hline

& \multicolumn{6}{c}{AUC} &\phantom{abc}& \multicolumn{6}{c}{FScore}& \phantom{abc} &\multicolumn{6}{c}{Accuracy}\\
\cmidrule{2-7} \cmidrule{9-14} \cmidrule{16-21}
Test dataset:   & 1&2&3&4&5& mean && 1&2&3&4&5& mean && 1&2&3&4&5& mean\\ 
\midrule
 \multicolumn{1}{c}{\textbf{window size: T=100}} \\ \hline
   LSTM, $\ell=1$\tnote{1}, N=256 & 71,78 &65,82&	96,67&	48,29&	53,77 & 66,87 &&	56,25&	65,10&	80,74&	52,71&	63,09 & 63,58&&	60,45&	55,26&	83,20&	43,73&	52,02  & 58,93\\ 
   FC1+FC2+LSTM, $\ell=1$,  N=256 &96,33&	\textbf{69,31} &	98,88&	94,63& 92,97 & \textbf{90,61} &&	86,75&	78,17&	\textbf{90,42}&	85,45&	89,64& 86,13 &&	89,48&	67,59&	\textbf{92,77}&	83,64&	85,53& 83,60 \\ 
   FC1+FC2+LSTM, $\ell=2$, N=128 & \textbf{96,51}&	66,98&	99,01&	94,20&	93,33 & 90,01 &&	\textbf{87,95}&	\textbf{80,42}&	89,52 &	86,10&	\textbf{89,96} & \textbf{86,79}&&	\textbf{90,83}&	\textbf{69,69} &	91,70&	83,35 &	\textbf{86,25} & \textbf{84,36}\\
   FC1+FC2+LSTM, $\ell=2$, N=256 & 96,36 &	67,50 &	\textbf{99,08} &	\textbf{94,69} &	\textbf{93,79} & 90,28 &&	87,71 &	79,05 &	89,76&	\textbf{86,67} &	89,21 & 86,48 &&	90,51&	68,24 &	91,92 &	\textbf{83,90} &	85,44 & 84,00\\ \hline

     \bottomrule
\end{tabular}}
   \begin{tablenotes}
     \item[1] $\ell$ denotes the layers
   \end{tablenotes}
  \end{threeparttable}

\caption{Exploration of the sequnce window size effect}\label{tab:lstms_window}
\centering
    \renewcommand*{\arraystretch}{}
    \resizebox{\textwidth}{!}{%
    \large
\begin{tabular}{@{}lcccccccccccccccccccc@{}}\toprule
\hline
 \multicolumn{1}{c}{\textbf{FC1+FC2+LSTM, $\ell=2$, N=128}} \\ \hline
& \multicolumn{6}{c}{AUC} &\phantom{abc}& \multicolumn{6}{c}{FScore}& \phantom{abc} &\multicolumn{6}{c}{Accuracy}\\
\cmidrule{2-7} \cmidrule{9-14} \cmidrule{16-21}
 Test dataset  & 1&2&3&4&5& mean && 1&2&3&4&5& mean && 1&2&3&4&5& mean\\ 
\midrule

   window size: T=50 & \bf{97,26}&63,46& 98,79 &	93,90 &	92,78 & 89,24 &&	86,97&	77,74&	88,07&	\textbf{87,84}&	88,90 &85,09 &&	89,43&	66,35 &	90,38&	\textbf{85,13}&	85,03 & 83,26\\
   window size: T=100 & 96,51&	66,98&	\textbf{99,01}&	\textbf{94,20}&	\textbf{93,33} & \textbf{90,01} &&	\textbf{87,95}&	\textbf{80,42}&	89,52 &	86,10&	\textbf{89,96} & \textbf{86,79}&&	\textbf{90,83}&	\textbf{69,69} &	91,70&	83,35 &	\textbf{86,25} & \textbf{84,36}\\
  window size: T=200 & 96,11	&\textbf{68,49}	&98,99	&91,99	&93,02 & 89,72	&&86,56	&79,57	&\textbf{90,33}&	83,06&	89,50 & 85,80 &&	89,53& 68,80 &	\textbf{92,42}&	79,84&	85,53 & 83,22 \\
 \hline  
     \bottomrule
\end{tabular}}
\vspace{-.45cm}

\end{table*}
\section{Experimental Analysis and Results}
\subsection{Experimental Setup}
\noindent \textbf{Data Collection:}  The data used in this work were collected in  Agaplesion Bethanien Hospital - Geriatric Center with the participation of real patients. The participants presented moderate to mild mobility impairment, according to clinical evaluation. Table \ref{tab:demograph} presents the participants' demographics. The subjects walked with physical support of a passive robotic rollator, used for the purpose of data collection, in a specific hospital area,  while wearing a set of MoCap markers. The visual markers data are used for ground truth extraction. The data for the Gait Stability Prediction were provided by a Kinect v1 camera for capturing the upper body, and a Hokuyo rapid LRF for detecting the legs, which were mounted on the rollator (Fig. \ref{fig:motivation}).
 
\noindent \textbf{Ground truth extraction from visual markers:} For extracting the GT labels of the Gait Stability Prediction framework, we followed the methodology described in \cite{LUGADE2011406} for computing the CoM from visual markers. In our previous work \cite{gchalROMAN} we have thoroughly described the process of detecting the gait phases from the planar foot impact. As explained in Sec. \ref{gaitStabdescr} for specifying the body stability we have to analyze the CoM-BoS interaction. Therefore, we detect the BoS according to the respective gait phase and measure the distance of the planar CoM position and evaluate the respective margins of stability. Given the respective average measures of the stability margins found in literature \cite{LUGADE2011406}, we label each walking instance as Safe or Fall Risk. Those two states constitute the ground truth labels used for training the proposed network. 
We have, also, conducted statistical analysis on the CoM motion for tuning the UKF used in our framework. Further, we found the statistics of the BoS size and position w.r.t. the users' tibia (the level at which the LRF scans the user's legs) which were used for a baseline rule-based method described below.

\noindent \textbf{Dataset statistics \& Data Augmentation:}
We use a dataset containing data from five patients, resulting in about 11000 frames corresponding to more than 300sec of walking. The dataset consists of 73\% safe states, leading to a largely unbalanced dataset. To avoid overfitting, we applied data augmentation, by applying on the GT CoM data of the Safe states additive random noise exploiting the statistics extracted for the CoM position by initial experimentation, i.e. zero mean and standard deviations $\sigma_x=0,15 cm$ in the x-direction and  $\sigma_y=0,20 cm$ in the y-direction, thus increasing the chances of detecting Fall Risk states. The same noise vector was also applied on the pose-based CoM. The final dataset consists of about 22000 frames, with about 46\% Fall-Risk labels. We employ a leave-one-out strategy for training/testing, i.e. data from four subjects in training and one in testing iteratively for cross-validation.

\subsection{Evaluation Strategy}
\noindent \textbf{Evaluation Metrics:} For the evaluation of the predicted gait stability labels, we present a thorough analysis regarding different LSTM architectures and other baseline methods by reporting the FScore, Accuracy, Precision and Recall metrics. We also evaluate the Area Under Curve (AUC), which is defined as the area under the Receiver Operating Characteristic (ROC) curve \cite{davis2006relationship,KOUTRAS201515}. 
The ROC curve relates true positive rates w.r.t. false positive rates at different classification thresholds, while AUC helps evaluating the classifier's performance across all possible classification thresholds.
We also demonstrate a brief validation of the CoM position estimation w.r.t. the respective extracted ground truth CoM employing the Root Mean Square Error (RMSE). 

\noindent \textbf{Baseline Methods:} As baseline methods we use the nonlinear Support Vector Machine (SVM) classifier with Gaussian kernel and the same inputs as the LSTM-based method, as well as a rule-based approach. The rule-based approach follows the biomechanics rules also employed for the ground truth extraction. From the augmented gait state estimation we get the leg's positions (at knee height) and the respective gait phase. We define the BoS in the horizontal plane employing the statistics learned by the visual markers analysis. Subsequently, we apply the same thresholds for evaluating the stability margins, which were used in the ground truth labels extraction.

\noindent \textbf{Implementation:}
We trained from scratch the proposed network with $N=128$ hidden layers in PyTorch using an Nvidia Titan X GPU. The output units of the FC layers FC1, FC2 were $4 \cdot N$ and $2 \cdot N$ respectively. For training we employed the Adam optimizer with the Logistic Cross-Entropy Loss for binary classification:

\vspace{-0.15cm}
{\small
\begin{equation}
\begin{multlined}
\mathcal{L}\left(\mathbf{W}, \mathbf{b}\right)
= -\sum_{j \in Y_+} \log{P\left(y_t^j=1|\mathbf{Z};\mathbf{W}, \mathbf{b}\right)} \\
-\sum_{j \in Y_-} \log{P\left(y_t^j=0|\mathbf{Z};\mathbf{W}, \mathbf{b}\right)} \label{eq:cross_entropy}
\end{multlined}
\end{equation}}

\vspace{-0.15cm}
where $\mathbf{Z}$ are the batched training samples, $y_t^j \in \{0,1\}$ is the binary gait stability label of $\mathbf{Z}$, and $Y_+$ and $Y_-$ are the safe and fall risk labelled sample sets respectively. $P(\cdot)$ is obtained by the softmax activation of the final layer (Eq. \ref{eq:softmax}). We used mini-batches of 256 clips, with initial learning rate $0.001$, momentum $0.9$ and weight decay $0.01$. The learning rate is divided by 10 after half of the epochs. For better regularization and faster training we also have applied dropout with probability $0.7$ in the FC1 layer and batch normalization \cite{Ioffe2015} after the first two FC layers and the LSTM unit. All subsystems were integrated in ROS \cite{ROS}.

\subsection{Experimental Results}
   
\noindent \textbf{{Validation of CoM estimation:} }
Table \ref{tab:com_rmse} reports the average position RMSE for the CoM w.r.t. the ground truth CoM positions extracted by the visual markers. Only subjects $\#$2 and $\#$4 present higher errors with RMSE 6cm. Fig. \ref{fig:CoM_estim1}  depicts the forward CoM displacement, as this is estimated by the UKF (blue line) given the detected CoM from the pose (orange line) w.r.t. the extracted ground truth (red line). We aim to show the difficulty of the dataset, since the ambiguity in the proximal pose detection leads to many occlusions and misdetections which are handled well by the UKF approach.

\noindent\textbf{Evaluation of the LSTM-based prediction model:}
Table \ref{tab:lstms_exam} explores the performance of different network  architectures for fixed window size sequences of 100 instances across the whole dataset using the metrics AUC, FScore and Accuracy. We observe that the plain architcture of a simple 1-layer LSTM with N=256 hidden states cannot capture the complexity of the body motion and decode the underlying interaction of the CoM with the legs motion per gait phase. On the contrary, the other models that use the two fully connected layers FC1 and FC2 before the LSTM cells can encode the hidden parameters of the whole body motion, achieving high accuracy. More importantly all three architectures achieve high AUC over 90\%, while the FScores are over 86\% meaning that they can be used for providing precise predictions about the safety of the rollator user. As for per user performance, it is evident that all network architectures perform worse when predicting subject's $\#2$ stability state. This poor performance is highly influenced by the higher CoM estimation errors as can be seen in Table \ref{tab:com_rmse}, or even a kind of pathological gait unseen to the system, since we only had available data from five patients. However, the existence of  patient $\#2$ data in the other four training datasets did not influence the performance; indeed the deep networks could decode the different types of walking and achieve high performance in spite of the estimation noise and errors from the tracking systems. 
  
From this exploratory study, we choose as our main network architecture the FC1+FC2+LSTMs with 2 layers of N=128 hidden variables, as it achieves the best FScore and Accuracy rate, while hitting a 90\% mean AUC. In Table \ref{tab:lstms_window}, we explore the influence of the temporal window size on the proposed model. We cross-evaluate the results for all datasets and present the AUC, FScore and Accuracy measures. We observe that the most consistent performance across all metrics is the one for input sequences of 100 instances.

\begin{table}
\caption{Comparison with baseline methods}\label{tab:resALL}
\centering
\begin{adjustbox}{width=0.5\textwidth}
\begin{tabular}{l|ccccccc}\toprule
\hline
\backslashbox{Metric}{Method} & Rule-based &\phantom{abc}& SVM& \phantom{abc} & LSTM & \phantom{abc}& FC1+FC2+LSTM \\ \hline
\midrule
   Precision& 83,27  && 80,59&& 66,87 &&\textbf{88,84}\\ 
   Recall& 77,29&& \textbf{88,11}&& 64,36 &&85,65 \\
   FScore& 80,09 && 83,22&& 63,58 && \textbf{86,79}  \\ 
   Accuracy& 80,81 &&80,62&& 67,27 &&\textbf{84,36}\\
   AUC& - && 86,73&& 58,93 &&\textbf{90,01} \\
     \bottomrule
\end{tabular}
\end{adjustbox}
\vspace{-0.5cm}
\end{table}

The final part of the experimental evaluation presents the cross-examination of the proposed model w.r.t. baseline methods. As described previously, we compare the LSTM-based network with the rule-based method, an SVM classifier and for the sake of generality with the simple LSTM (no initial FC layers) of Table \ref{tab:lstms_exam}. We evaluate the average metrics across all datasets. Inspecting the results, the rule based method achieves an accuracy of 80\% and an analogous FScore. This finding is a strong indicator that our analysis about predicting the stability state by fusing the data from the body pose and the gait state estimates is plausible. The rule-based method is a discrete process that does not include probabilities computation like the rest of the methods; thus the AUC could not be evaluated. 

On the other hand, we notice that the SVM classifier performs very well. This was well expected, since the SVM classifier with nonlinear kernel is known to work well in binary classification problems with relatively small datasets. However, the proposed network improves the SVM scores as it achieves about 4\%  better FScore and ameliorates the AUC score by approximately 3\%. The plain LSTM model achieves again the poorest results, as in Table \ref{tab:lstms_exam}, since it is evident that there is an underlying nonlinear relation between the CoM and the gait states, which is encoded by the fully connected layers. For better understanding of the results of this table, we plot the ROC curve for the SVM, LSTM, the proposed network, and the random predictor across all possible classification thresholds. Although, all methods perform better than random predictor, the proposed fully connected LSTM-based network outperform all the other methods at all cases. 

\begin{figure}
      \centering
      \includegraphics[width=0.35\textwidth,trim={1.2cm 0cm 1.2cm 0cm},clip]{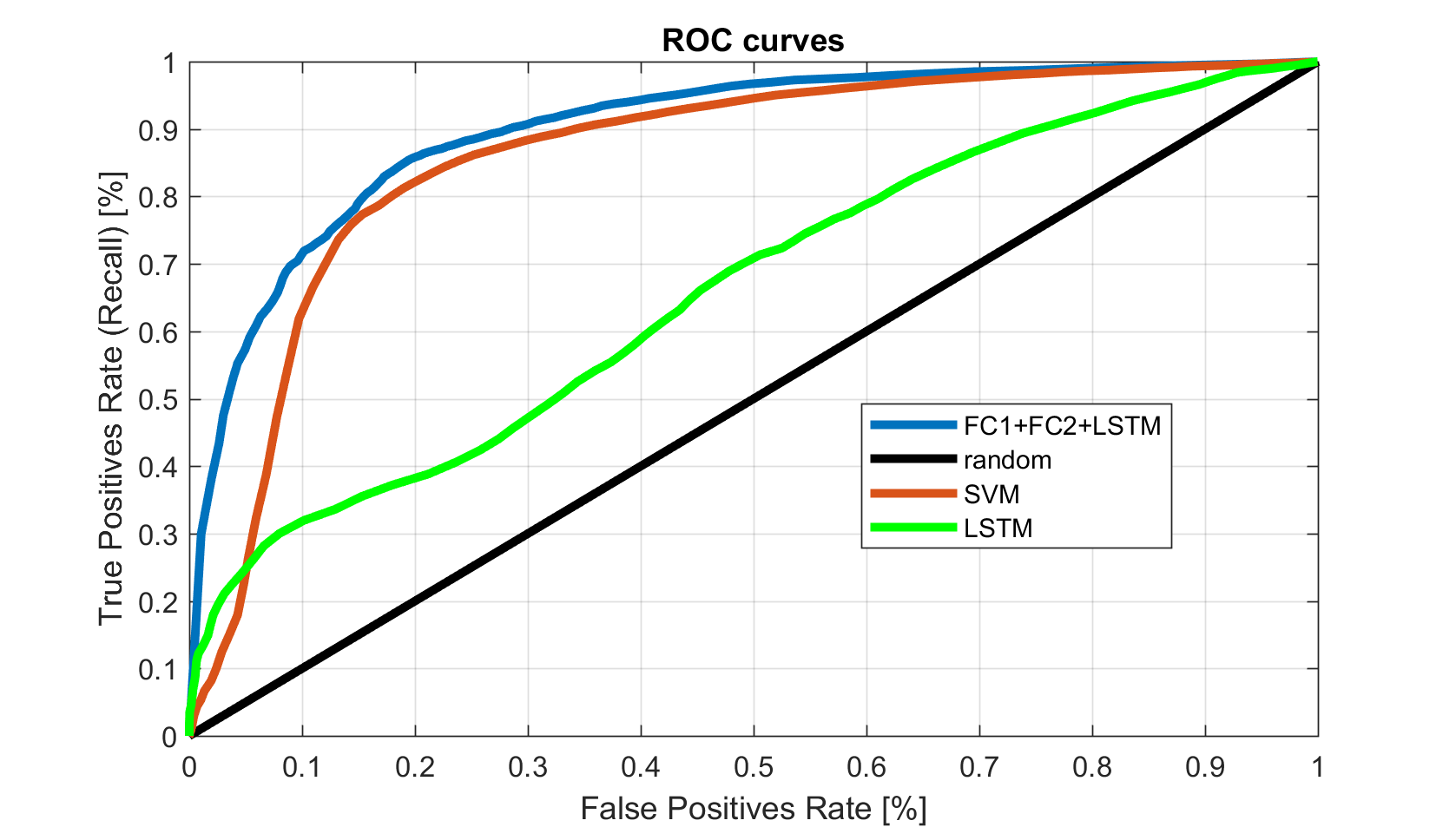}
      \caption{\small ROC curve of the SVM, LSTM and fully connected LSTM along with the chance predictor.}       
      \label{fig:roc}
      \vspace{-0.55cm}
\end{figure}

\section{Conclusions \& Future Work}
In this work, we have presented and experimentally evaluated a novel and robust method for on-line gait stability analysis of elderly subjects walking with a mobility assistant platform. The proposed method is fusing the information from an RGB-D camera which captures the upper body and a LRF which monitors the legs motion. We use the OP framework for detecting the upper body pose, we track the respective CoM, while we exploit the LRF data in an augmented gait state estimation framework for extracting the legs' position and the respective gait phase. Our main contribution is the  proposal of a novel LSTM-based network for predicting the stability state of the elderly, by classifying walking as safe or fall-risk at each instant. 

In the future, we plan to increase our datasets for training and evaluation of the proposed model and integrate it into a user-adaptive context-aware robot control architecture with a fall-prevention functionality for an intelligent robotic assistant platform.
%

%
\clearpage
\bibliography{Ref_iWalk_ICRA19}
\bibliographystyle{IEEEtran}

\end{document}